\documentclass[10pt,twocolumn,letterpaper]{article}

\usepackage{cvpr}
\usepackage{times}
\usepackage{epsfig}
\usepackage{graphicx}
\usepackage{amsmath}
\usepackage{amssymb}
\usepackage{bm}
\usepackage{multirow}
\usepackage{url}
\usepackage{float}
\usepackage[numbers,square]{natbib}
\usepackage[breaklinks=true,bookmarks=false]{hyperref}

\cvprfinalcopy 

\begin{document}

\title{Swin-X2S: Reconstructing 3D Shape from 2D Biplanar X-ray \\ with Swin Transformers}

\author{
Kuan Liu\footnotemark[1]\ \ , Zongyuan Ying\footnotemark[1]\ \ , Jie Jin\footnotemark[1]\ \ , Dongyan Li\footnotemark[1]\ \ , Ping Huang\footnotemark[1]\ \ , Wenjian Wu\footnotemark[1]\ \ , Zhe Chen\footnotemark[1]\ \ ,\\ Jin Qi\footnotemark[1]\ \ , Yong Lu\footnotemark[2]\ \ , Lianfu Deng\footnotemark[1]\ \ , and Bo Chen\footnotemark[1]\ \\\
}

\maketitle
\renewcommand{\thefootnote}{\fnsymbol{footnote}}
\footnotetext[1]{Department of Orthopaedics, Shanghai Key Laboratory for Prevention and Treatment of Bone and Joint Diseases, Shanghai Institute of Traumatology and Orthopaedics, Ruijin Hospital, Shanghai Jiao Tong University School of Medicine, Shanghai 200025, China.}
\footnotetext[2]{Department of Radiology, Ruijin Hospital Luwan Branch, School of Medicine, Shanghai Jiaotong University,Shanghai 200003, China.}
\renewcommand{\thefootnote}{\arabic{footnote}}

\begin{abstract}
	\setlength{\baselineskip}{1em}
The conversion from 2D X-ray to 3D shape holds significant potential for improving diagnostic efficiency and safety. However, existing reconstruction methods often rely on hand-crafted features, manual intervention, and prior knowledge, resulting in unstable shape errors and additional processing costs. In this paper, we introduce Swin-X2S, an end-to-end deep learning method for directly reconstructing 3D segmentation and labeling from 2D biplanar orthogonal X-ray images. Swin-X2S employs an encoder-decoder architecture: the encoder leverages 2D Swin Transformer for X-ray information extraction, while the decoder employs 3D convolution with cross-attention to integrate structural features from orthogonal views. A dimension-expanding module is introduced to bridge the encoder and decoder, ensuring a smooth conversion from 2D pixels to 3D voxels. We evaluate proposed method through extensive qualitative and quantitative experiments across nine publicly available datasets covering four anatomies (femur, hip, spine, and rib), with a total of 54 categories. Significant improvements over previous methods have been observed not only in the segmentation and labeling metrics but also in the clinically relevant parameters that are of primary concern in practical applications, which demonstrates the promise of Swin-X2S to provide an effective option for anatomical shape reconstruction in clinical scenarios. Code implementation is available at: \url{https://github.com/liukuan5625/Swin-X2S}.
\end{abstract}

\section{Introduction}
X-ray imaging is an essential tool for preliminary clinical assessments and is widely used in emergency care, orthopedics, dentistry, and routine health check-ups due to its cost-effectiveness, low radiation exposure, and rapid imaging capabilities. However, its inherent 2D structure limits its ability to distinguish spatial shapes and intricate anatomical details. In contrast, CT scans offer high-resolution 3D representations, providing comprehensive views of bones, organs, blood vessels, and soft tissues, making them ideal for detecting subtle abnormalities. But CT scans are significantly more costly, entail higher radiation exposure, and require longer processing time. 

The challenge of reconstructing 3D shape from 2D X-ray images has attracted long-lasting attention due to its potential capability to provide the interest anatomical structures for preliminary diagnosis and treatment in an economical, safe, and rapid manner \cite{Brown1976, Herman2009, Hosseinian2015, Ying2019, Shakya2024}. Especially in poverty-stricken areas where access to CT scanners is limited or unavailable \cite{Hricak2021, Ngoya2016}, and for underdeveloped children who must undergo CT radiation exposure, such as those with scoliosis or fractures \cite{Aubert2019, Illes2012}, the importance of 2D-3D reconstruction becomes even more pronounced to these vulnerable populations. 

Early approaches to 3D shape reconstruction relied on hand-crafted features, including point-based, contour-based, statistical shape model (SSM) based, and active shape model (ASM) based algorithms. These methods require manual intervention and prior knowledge of anatomical geometry, leading to unacceptable shape error and processing time in clinical practice. With the development of deep learning, there has been a shift toward data-driven methods that automatically learn patterns from data. These approaches, such as convolution neural networks (CNNs) \cite{Krizhevsky2012, He2016} and Transformers \cite{Vaswani2017}, have shown promise in overcoming the limitations of traditional methods \cite{Maken2023}. Deep neural networks are capable of directly reconstructing 3D structures from X-ray images, thereby reducing the reliance on physician expertise and improving both accuracy and efficiency. Nevertheless, the effectiveness of deep learning algorithms heavily depends on the quality and diversity of annotated datasets, and existing methods typically design distinct algorithms for different anatomies, some of which require specific data processing methods \cite{Ying2019, Shakya2024}.

In this paper, we propose a novel end-to-end deep learning-based method called Swin-X2S for directly reconstructing 3D segmentation and labeling from biplanar raw X-ray images. The proposed algorithm employs a 2D transformer-based encoder and a 3D convolution-based decoder, with a dimension-expanding module to bridge 2D pixels and 3D voxels. To evaluate the performance of proposed method, we conduct thorough experiments across 9 publicly available datasets, covering 54 osseous structures from four anatomical regions (femur, hip, spine, and rib). We report segmentation and labeling results, along with shape errors in clinically relevant parameters.

\begin{figure*}[!t]
	\centering
	\includegraphics[width=1.0\textwidth]{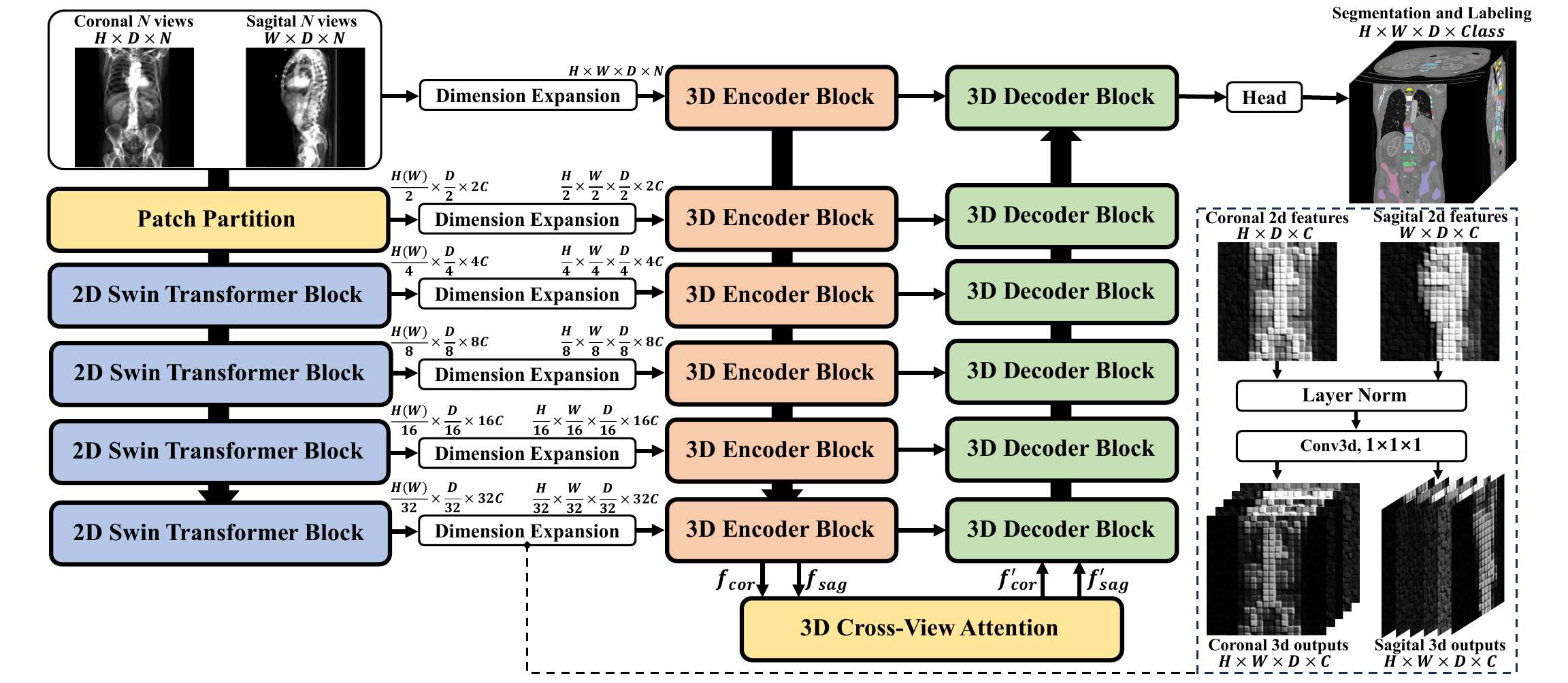}
	\caption{\small The overall architecture of the proposed Swin-X2S network for CT Segmentation and labeling from biplanar X-ray images (${N\text{=}1}$). Swin-X2S takes paired inputs: the coronal view and sagittal view X-ray images. The network generates 2D pyramid features via transformers, which are applied to dimension expansion modules (right dashed box) for upscaling and then reconstruct through 3D U-shaped convolution network.}
	\label{fig01}
\end{figure*}

\section{Related work}
Reconstructing 3D structures from 2D projections is an inverse problem that is inherently ill-posed and computationally challenging. Moreover, the arbitrary field of view (FOV), the distortion of X-ray projection, the presence of related diseases and metal implants introduce further challenges. To address above problems, existing studies resort to non-learning methods align with prior knowledge and data-driven deep learning approaches.

Traditional methods rely on a combination of anatomical prior knowledge, geometric landmarks and statistical features. Early works identified spatial landmarks of each vertebra from biplanar X-rays to determine the configuration of vertebral column \cite{Brown1976, Mitton2000, Mitulescu2002}. In order to reduce human intervention and reconstruction time, semi-automated statistical methods \cite{Benameur2003, Herman2009, Boisvert2011, Whitmarsh2013, Karade2015, Anas2016} have been proposed that utilize hand-crafted features (points, contours, atlases or deformation fields) to construct SSMs or ASMs for reconstructing various anatomical regions (vertebrae, hips, femurs, wrists). Despite incorporating statistical inferences, these methods still require manual pre-processing or post-processing steps to handle complex morphologies, which introduce extra human error and processing cost.

As deep learning methods have gained wider adoption in medical image processing, researchers have proposed data-driven algorithms to reduce user supervision. Some of these methods aim to develop neural networks incorporate with traditional method to improve feature extraction and model efficiency \cite{Aubert2019, Chenes2021, Song2021}. For example, Aubert et al. \cite{Aubert2019} utilized CNNs to develop a realistic SSM of spine from biplanar X-ray images for automatic vertebral column reconstruction; Oral-3D \cite{Song2021} reconstructed 3D structure of oral cavity based on prior information of the dental arch from a single panoramic X-ray. 

At the same time, scholars also have proposed purely deep learning algorithms that eliminate the reliance on hand-crafted features and instead learn shape distributions directly from the training data. Some approaches first reconstruct raw CT images from X-ray projections and then perform 3D segmentation on the reconstructed CT \cite{Ying2019, Shen2019, Ge2022, Kyung2023}. For instance, X2CT-GAN \cite{Ying2019} was the first method proposed to reconstruct raw CT images from two orthogonal X-ray images in an end-to-end manner. Shen et al. \cite{Shen2019} introduced a deep learning framework to generate volumetric tomography from a single X-ray image. X-CTRSNet \cite{Ge2022} and PerX2CT \cite{Kyung2023} both adopt encoder-decoder architecture to further enhance the reconstruction performance. However, reconstructed raw CT usually suffers from blurry volumes and missing details, which introduces difficulties in subsequent segmentation tasks. 

Methods that directly reconstruct 3D shape from X-rays have demonstrated superior performance simultaneously. Kasten et al. \cite{Kasten2020} presented an end-to-end CNN approach for 3D shape reconstruction of knee bones directly from biplanar X-ray images. Several other studies \cite{Shiode2021, Almeida2021, Bayat2020, Nakao2021} have demonstrated the feasibility of algorithms for reconstructing anatomical structures such as the wrist, pelvis, vertebrae, and liver from corresponding X-ray images. For anatomical structures with strong similarities, such as vertebrae or ribs, different parts of overall morphology often appear highly alike, making them challenging to differentiate in X-ray images even for experienced radiologists. Most existing methods employ additional processes to address this challenge \cite{Nakao2021, Chen2023, Wu2023, Shakya2024}. BX2S-Net \cite{Chen2023} is able to segment each vertebra of spine with high accuracy, however, their method  requires known spatial location of each vertebra to crop raw image into patches, each containing individual vertebra images, before performing shape reconstruction. Wu et al.\cite{Wu2023} proposed a spine localization and identification method from multi-view projection images, which localizes the centroid of each vertebra but does not establish its shape, with labeling serves as a necessary condition for segmentation. 

These aforementioned methods generally require customized network designs and dataset processing pipelines tailored to specific structures, which limits their applicability. A recent study \cite{Shakya2024} introduced a benchmark of 3D bone shape reconstruction from biplanar X-ray images, performing four anatomies (femur, hip, spine and rib) 3D shape reconstruction across 6 public datasets. This work shares a similar objective with our study, yet it only segments binary shape masks, and notably, their method also necessitates precise localization of each vertebra to crop raw X-ray images, which serves as a prerequisite for spine reconstruction. To the best of our knowledge, our proposed framework is the first capable of accommodating these four anatomical structures and producing 3D segmentation and labeling directly from biplanar 2D X-ray images. 

\section{Method}
The particularity of 2D-3D reconstruction task lies in the mismatch dimensions of the input and output. As depicted in Fig. \ref{fig01}, the overall architecture of Swin-X2S can be divided into three steps: \textbf{(i)} 2D Swin-Transformer-based encoder takes paired raw X-ray images as input to facilitate image feature extraction; \textbf{(ii)} Dimension-expanding module shuttles low-level tokens from 2D encoder to 3D decoder; \textbf{(iii)} 3D convolution-based decoder incorporates a cross-attention mechanism to fully integrate information from biplanar views, thereby enabling the shape reconstruction of various anatomies.

\subsection{2D Swin Transformer encoder}
The inputs of Swin-X2S are ${\bm{\chi} _{cor}} \in { {\mathbb R} ^{H \times D \times N}}$ and ${\bm{\chi}_{sag}} \in { {\mathbb R} ^{W \times D \times N}}$, where $\left( {H \times W \times D} \right)$ represents the original CT scan size and ${N}$ denotes the number of X-ray images (see section \ref{drr}, ${N\text{=}1}$ denotes biplanar X-ray inputs). Embedded tokens are fed into successive Swin Transformer blocks \cite{Liu2021} to generate 2D pyramid features. At block ${l}$, 2D tokens are divided into ${N}$ non-overlapping ${M \times M}$ patches, each window patch conducts self-attention computation to establish pixel correlation. In the subsequent block ${l \text{+} 1}$, 2D images are shifted by $\left( {\frac{M}{2}, \frac{M}{2}} \right)$ pixels and then similarly divided into flattened windows to compute self-attention.  Consecutive Swin Transformer blocks switch back and forth between patch windows and shifted patch windows. Specifically, the calculation is as follows: 
\begin{equation}
	{\bm{\hat z}^{l}} = {\rm{W \text{-} MSA}}\left({{\rm{LN}}\left( {{\bm{z}^{l - 1}}} \right)} \right) + {\bm{z}^{l - 1}}
\end{equation}
\begin{equation}
	{\bm{z}^{l}} = {\rm{MLP}}\left({{\rm{LN}}\left( {\bm{\hat z}^{l}} \right)} \right) + {\bm{\hat z}^{l}}
\end{equation}
\begin{equation}
	{\bm{\hat z}^{l+1}} = {\rm{W \text{-} MSA}}\left({{\rm{LN}}\left( {{\bm{z}^{l}}} \right)} \right) + {\bm{z}^{l}}
\end{equation}
\begin{equation}
	{\bm{z}^{l+1}} = {\rm{MLP}}\left({{\rm{LN}}\left( {\bm{\hat z}^{l+1}} \right)} \right) + {\bm{\hat z}^{l+1}}
\end{equation}

W-MSA and SW-MSA are multi-head self-attention modules that alternate consecutively. The self-attention mechanism is formulated as:
\begin{equation}
	\bm{ATTN}(Q,K,V) = \rm{softmax} \left( {\frac{{\mathit{Q}{\mathit{K}^T}}}{{\sqrt d }} + \mathit{B}} \right)\mathit{V}
\end{equation}
Where ${\bm{Q},\bm{K},\bm{V}} \in { {\mathbb R} ^{H \times D \times N}}$ denote queries, keys, and values respectively; ${\bm{B}}$ represents the relative position of tokens within each window.

\subsection{Dimension expanding module}
The dimension transformation issue does not arise in pure 2D or 3D encoder-decoder networks, however, the mapping from 2D images to 3D structures is essential in shape reconstruction task. It's evident that coronal photograph lack structural information from the sagittal view, and vice versa. We propose a simple and effective dimension expanding module, as illustrated in Fig. \ref{fig01}, to convert 2D pixel features to 3D voxel features at each output of 2D Swin Transformer patch merging stage. To be more specific, the coronal image and sagittal image are stacked into the same resolution of ${{\mathbb R} ^{H \times W \times D \times C}}$ using $1\times1\times1$ 3D convolution kernels along sagittal and coronal axes, respectively. This alignment ensures that orthogonal 2D representations can be extracted and processed uniformly across different resolutions, enabling the integration of coherent and comprehensive information.

\subsection{3D U-shaped decoder}
3D Decoder adopt U-shaped network design \cite{Ronneberger2015, Hatamizadeh2021, Wang2023}. The contracted path processes high-dimensional tokens at each stage with dimensions of $\left( \frac{H}{2^i}, \frac{W}{2^i}, \frac{D}{2^i} \right)$ where $i = 0, 1, 2, 3, 4, 5$, hierarchically compressing the relevant essential features. The expansive path takes the outputs from the previous stage and concatenates them with the opposite side via skip connections, progressively upsamples and refines latent tokens to original resolution. The final segmentation masks are reconstructed by the classification head.

At the bottleneck of the U-shaped network $\left( {i = 5} \right)$, the cross-attention module is employed to integrate the high-level representations from different views. The fusion enables paired coronal view and sagittal view complement from each other to get spatial features which may be obstructed in original perspective. The cross-attention mechanism is formulated as follows: 
\begin{equation}
	{\bm {CrossATTN}}(Q_i,K_j,V_j) = \rm{softmax} \left( \frac{{\mathit{Q_i}{\mathit{K_j}^T}}}{{\sqrt d}}  \right)\mathit{V_j}
\end{equation}
In the equation, $i$ and $j$ represent different views, while $\bm{Q_i}$, $\bm{K_i}$ , and $\bm{V_j}$ correspond to the query, key, and value matrices, respectively.

\begin{table*}[!t]
	\fontsize{8}{10}\selectfont
	\centering
	\caption{\small The model hyperparameters of the Swin-X2S network, "Hidden dimension" denotes the channel number of the hidden layers.}
	\label{table01}
	\setlength{\tabcolsep}{5pt}
	\renewcommand\arraystretch{1.2}
	\begin{tabular}{ll|cccccc}
		\hline
		\multicolumn{2}{l|}{\multirow{1}{*}{Model Architecture \quad}} & \multicolumn{1}{l}{\multirow{1}{*}{\qquad Image Size \quad}} & \multirow{1}{*}{Hidden dimension \quad} & \multirow{1}{*}{Block numbers \quad} & \multirow{1}{*}{Head numbers \quad} & \multirow{1}{*}{Paramethers(M) \quad} & \multirow{1}{*}{FLOPs(T)} \\\hline
		\multicolumn{2}{l|}{Swin-X2S-Tiny} & {[}96,96,128{]} & 32 & {[}2,2,6,2{]} & {[}1,2,4,8{]} & 32.51 & 0.42 \\ \hline
		\multicolumn{2}{l|}{Swin-X2S-Small} & {[}128,128,160{]} & 64 & {[}2,2,6,2{]} & {[}2,4,8,16{]} & 129.97 & 3.35 \\ \hline
		\multicolumn{2}{l|}{Swin-X2S-Base} & {[}128,128,160{]} & 96 & {[}2,2,18,2{]} & {[}3,6,12,24{]} & 313.68 & 7.54 \\ \hline
		\multicolumn{2}{l|}{Swin-X2S-Large} & {[}128,128,160{]} & 128 & {[}2,2,18,2{]} & {[}4,8,16,32{]} & 557.58 & 13.38 \\ \hline
	\end{tabular}
\end{table*}

\subsection{Model architecture}
2D Swin Transformer is designed with four stages. A patch merging layer with a size of $2\times2$ is applied at the end of each stage to merge neighboring pixels. The 2D patch window size is set to $7\times7$ and the expansion ratio of MLP is set to $\alpha=4$. 3D Decoder block consists of two successive $3\times3\times3$ convolution layers with residual connection, scaling up the channel and scaling down resolution by a factor of 2 during feature extraction, and operating the opposite way during feature aggregation. We introduced four types of our model: Swin-X2S-Tiny, Swin-X2S-Small, Swin-X2S-Base, and Swin-X2S-Large. The embedding dimension is set to 32, 64, 96, 128 respectively. The input size, output size, model size, theoretical computational complexity (FLOPs), and throughput of the model variants for CT scan are listed in Table \ref{table01}. The model sizes of Swin-X2S-Tiny, Swin-X2S-Small, Swin-X2S-Large are approximately 0.1x, 0.5x, and 2x that of Swin-X2S-Base, respectively.

\begin{figure}[!b]
	\centering
	\includegraphics[width=0.5\textwidth]{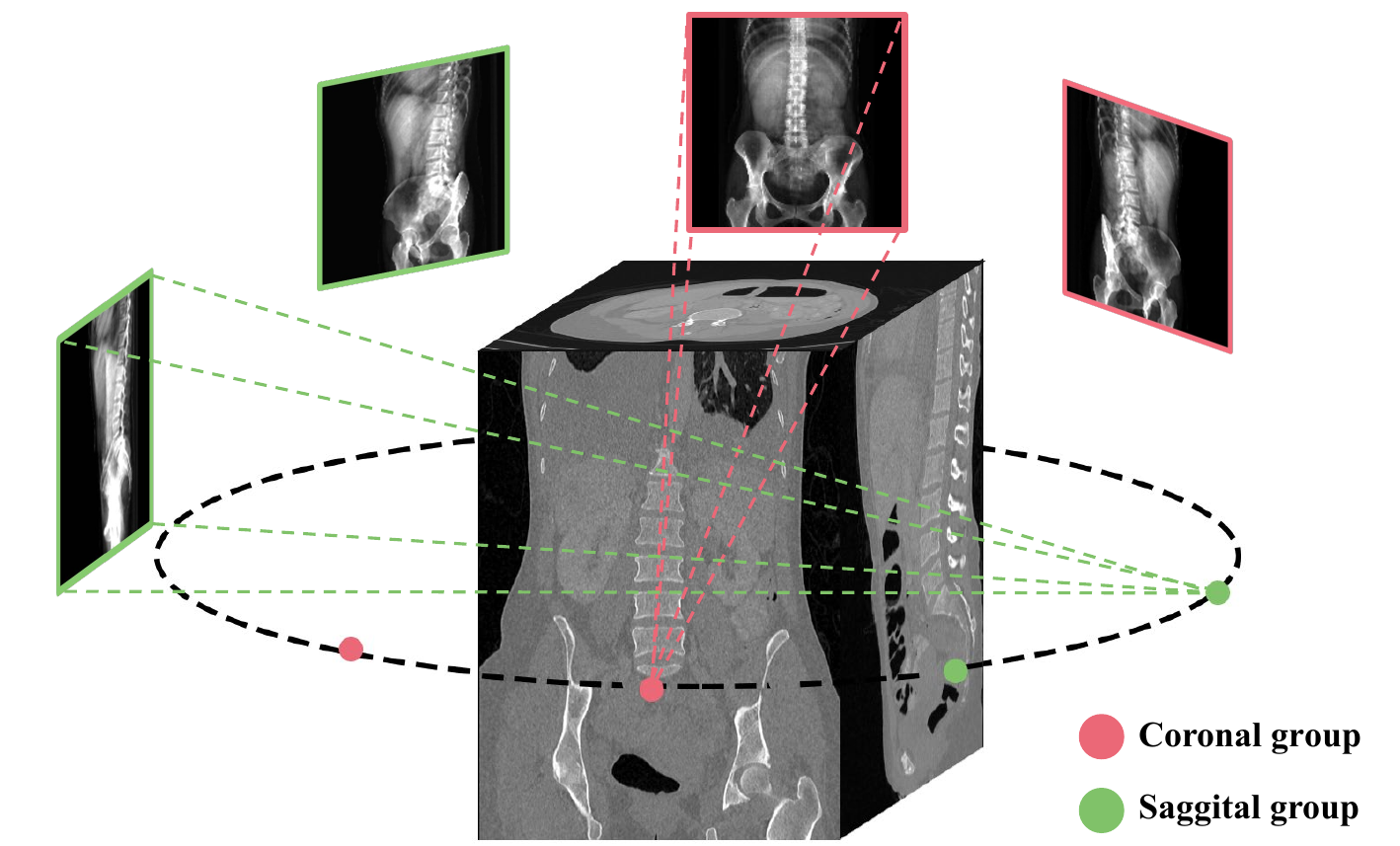}
	\caption{\small An example of generated coronal group and sagittal group based on DRR method, where the number of projection views $N$ is set to 2.}
	\label{fig02}
\end{figure}

\subsection{Loss function}
The overall loss function of Swin-X2S consists two parts: one measures the similarity between single view predictions and ground truth, the other computes the difference between predictions from different view:
\begin{equation}
	\begin{split}
		\mathcal L({p_{i}},{q_{i}},{y_{i}}) = \;\frac{1}{2}\;(&{\mathcal L_{single}}({p_{i}},{y_{i}}) + {\mathcal L_{cross}}({p_{i}},{q_{i}})\; + \\ &{\mathcal L_{single}}({q_{i}},{y_{i}}) + {\mathcal L_{cross}}({q_{i}},{p_{i}}))	
	\end{split}
\end{equation}
where $p$, $q$ and $y$ represent the coronal predicted mask, sagittal predicted mask and ground truth label, respectively.

For single view loss part, we adopt DiceCE loss \cite{Tang2022}, which combines Dice Loss \cite{Milletari2016} and Cross Entropy Loss \cite{Shannon1948} to measure segmentation performance in a supervised manner; For cross view loss part, we adopt KL divergence loss \cite{Zhang2018} to quantify the difference between view in an unsupervised manner.
\begin{equation}
	{\mathcal L_{single}} = 1 - \frac{1}{N}\sum\limits_{i = 1}^N {(\frac{2\left| {{p_i} \cap {y_i}}\right|}{\left| {{p_i}} \right| + \left| {{y_i}} \right|} + {{y_i}\log ({p_i})})}
\end{equation}
\begin{equation}
	{\mathcal L_{cross}} = \sum\limits_{i = 1}^N{{p_i}\log \frac{p_i}{q_i}}
\end{equation}
while $i$ denotes the $i$-th voxel of the total of $N$ CT voxels.

\section{Experimental setup}
\subsection{DRR generation}
\label{drr}
To evaluate our proposed method, a large dataset with paired biplanar images, corresponding CT slices, and segmentation masks is required. Furthermore, precise spatial alignment between CT and corresponding X-ray images is essential for accurate 3D reconstruction. 

Due to the lack of such a dataset, most existing methods have applied digitally reconstructed radiograph (DRR) technique \cite{Milickovic2000}, a ray casting approach in which virtual rays are projected through 3D volumetric data to synthesize 2D radiographic images that mimicking conventional X-ray. Similarly, we use DRR technology to generate coronal and sagittal view groups by rotating ray source, producing ${N}$ different view images at intervals of ${90/N}$ degrees from CT coronal and sagittal plane respectively, as shown in Fig. \ref{fig02}. Specifically, ${N\text{=}1}$ corresponds to orthogonal biplanar X-ray images. Based on the aforementioned methods, the original dataset used for 3D medical image segmentation has been restructured into a new dataset consisting of paired X-ray scans and CT masks for reconstructing 3D shapes from 2D images.

\subsection{Dataset}
We have collected five large-scale public CT segmentation datasets to comprehensively evaluate the performance of our algorithm: \textbf{(i)Totalsegmentator}\footnote{\url{https://github.com/StanfordMIMI/TotalSegmentatorV2}} \cite{Wasserthal2023} consists of 1228 CT scans and comprehensive labeling for multiple anatomical structures.  
\textbf{(ii)CTPelvic1K}\footnote{\url{https://github.com/MIRACLE-Center/CTPelvic1K}} \cite{Liu2021} comprises 1184 CT volumes, each with annotated sacrum, left hip, and right hip masks. \textbf{(iii)CTSpine1K}\footnote{\url{https://github.com/MIRACLE-Center/CTSpine1K}} \cite{Deng2021} contains 1005 CT volumes with labeled vertebrae masks. \textbf{(iv)VerSe'19} (Vertebral Segmentation Challenge 2019) \footnote{\url{https://github.com/anjany/verse}} \cite{Sekuboyina2021} comprises 160 spinal CT scans with ground truth annotations.
\textbf{(v)RibSeg v2}\footnote{\url{https://github.com/M3DV/RibSeg}} \cite{Jin2023} contains 660 CT scans with binary rib annotations.
These datasets were further subdivided into 9 subsets based on different anatomical structures, all of these subsets, with each subset following the official data split settings:

\noindent \textbf{TotalSegmentator-Femur-Dataset} 571 CT scans with femur masks were selected from the original TotalSegmentator dataset. This subset necessitates the segmentation of both the left and right femur.

\noindent \textbf{TotalSegmentator-Pelvic-Dataset} 1086 CT scans with pelvis masks were selected from the original TotalSegmentator dataset. This subset necessitates the segmentation of sacrum, left pelvis and right pelvis.

\noindent \textbf{CTPelvic1K-Pelvic-Dataset} CTPelvic1K consists of 1106 pelvis CT scans, this dataset necessitates the segmentation of 3 categories.

\noindent \textbf{TotalSegmentator-Spine-Dataset} 1189 CT scans with spine masks were selected from the original TotalSegmentator dataset. Angiography and overcropped scans were removed due to lack of context information. This subset necessitates the segmentation of 25 vertebrae categories.

\noindent \textbf{CTSpine1K-Spine-Dataset} CTSpine1K consists of 1005 spine CT scans, this dataset requires the segmentation of 25 categories.

\noindent \textbf{VerSe'19-Spine-Dataset} VerSe'19 consists of 160 spine CT scans, this dataset necessitates the segmentation of 25 classes. 

\noindent \textbf{TotalSegmentator-Rib-Dataset} 1086 CT scans with rib masks were selected from the original TotalSegmentator dataset. Samples with angiocardiography or containing too few ribs were excluded. This subset necessitates the segmentation of twelve pairs of ribs, totaling 24 categories.

\noindent \textbf{RibSeg v2-Rib-Dataset} RibSeg v2 contains 660 CT scans, with most cases are confirmed with complete rib cages. Binary rib masks are re-assigned into 24 rib classes based on region connectivity.

\noindent \textbf{TotalSegmentator-All-Dataset} This dataset is the union of all the previous TotalSegmentator subsets, totaling 1005 CT scans, necessitates the segmentation of 4 anatomies comprising 54 categories.

\subsection{Evaluation metrics}
\subsubsection{Segmentation and labeling metrics}
We adopt segmentation and labeling metrics to evaluate the performance of reconstruction \cite{Sekuboyina2021}. Segmentation metrics: (i) Dice Coefficient (Dice), a voxel overlap-based metric measures the quality of the segmentation which ranges from $0\%$ (zero overlap) to $100\%$ (perfect overlap). (ii) Hausdorff Surface Distance 95 (HD), distance-based metric measures the local maximum distance between the surface of prediction mask and the surface of ground truth; Labeling metrics: (i) Localization error (L-Error), a distance-based metric measures the average Euclidean distance between the ground truth centroids and the predicted segmentation centroids. (ii) Identification rate (ID-rate), accuracy-based metric measures the proportion of correctly identified instances out of the total number of target instances.

\subsubsection{Clinically Relevant Metrics}
For practical applications of 3D reconstruction, depending on the downstream task, clinical experts are more interested in the specific morphological structures of certain anatomical regions rather than overly general statistical segmentation or labeling metrics \cite{Kofler2021, Shakya2024}. Error assessment of morphological parameters for specific anatomies is a crucial step toward the practical application of 2D-3D reconstruction. We adopt automated clinical parameters evaluation pipelines to comprehensively analyze each sample: \textbf{Femur Morphometry} We adopt the method from \cite{Cerveri2010} to measure Femoral Head Radius (FHR) and Neck Shaft Angle (NSA) from femur segmentation. \textbf{Pelvic Landmarks} Fischer et al. \cite{Fischer2019} automatically recognizes bony landmarks of the pelvis. We measure the landmark distance between prediction and ground truth. \textbf{Vertebra Morphometry} Di Angelo and Di Stefano \cite{Di2015} proposed an automatic method to identify geometric references and the associated dimensions. We employ this method to measure the morphological features of each vertebra in the reconstructed spine. \textbf{Rib Centerline} Jin el al. \cite{Jin2023} emphasized the clinical importance of rib centerline. Following their pipeline, we automatically perform rib centerline extraction and evaluation.

\begin{table}[!t]
	\fontsize{8}{10}\selectfont
	\centering
	\caption{\small Quantitative evaluation results of the 9 subsets using our proposed Swin-X2S-Base network with biplanar inputs. Dice score (Dice) and Hausdorff surface distance (HD) for segmentation evaluation, localization error (L-Error) and identification rate (ID-rate) for labeling evaluation.}
	\label{table02}
	\setlength{\tabcolsep}{3pt}
	\renewcommand\arraystretch{1.2}
	\begin{tabular}{l|cc|cc}
		\hline
		\multicolumn{1}{l|}{\multirow{2}{*}{Dataset}} & \multicolumn{2}{c}{\multirow{1}{*}{Segmentation results}} & \multicolumn{2}{c}{\multirow{1}{*}{Labeling results}} \\ \cline{2-5}
		\multicolumn{1}{l|}{} & \multirow{1}{*}{Dice(\%)↑} & \multirow{1}{*}{HD(mm)↓} & \multirow{1}{*}{L-error(mm)↓} & \multirow{1}{*}{ID-rate(\%)↑} \\ \hline
		
		\multirow{1}{*}{TotalSeg-Femur} & 92.31 & 4.79 & 2.79 & 99.07 \\ \hline
		\multirow{1}{*}{TotalSeg-Pelvic} & 86.47 & 5.24 & 3.03 & 98.21 \\ \hline
		\multirow{1}{*}{CTPelvic1K-Pelvic} & 89.11 & 4.66 & 2.57 & 98.79 \\ \hline
		\multirow{1}{*}{TotalSeg-Spine} & 75.21 & 6.76 & 3.39 & 90.57 \\ \hline
		\multirow{1}{*}{CTSpine1K-Spine} & 83.84 & 5.27 & 2.79 & 87.76 \\ \hline
		\multirow{1}{*}{VerSe'19-Spine} & 64.54 & 10.77 & 7.43 & 77.23 \\ \hline
		\multirow{1}{*}{TotalSeg-Rib} & 45.92 & 16.41 & 13.27 & 78.36 \\ \hline
		\multirow{1}{*}{RibSeg v2-Rib} & 56.43 & 13.42 & 11.20 & 83.32 \\ \hline
		\multirow{1}{*}{TotalSeg-All} & 57.51 & 13.33 & 9.22 & 83.56 \\ \hline
	\end{tabular}
\end{table}

\begin{figure*}[!t]
	\centering
	\includegraphics[width=1\textwidth]{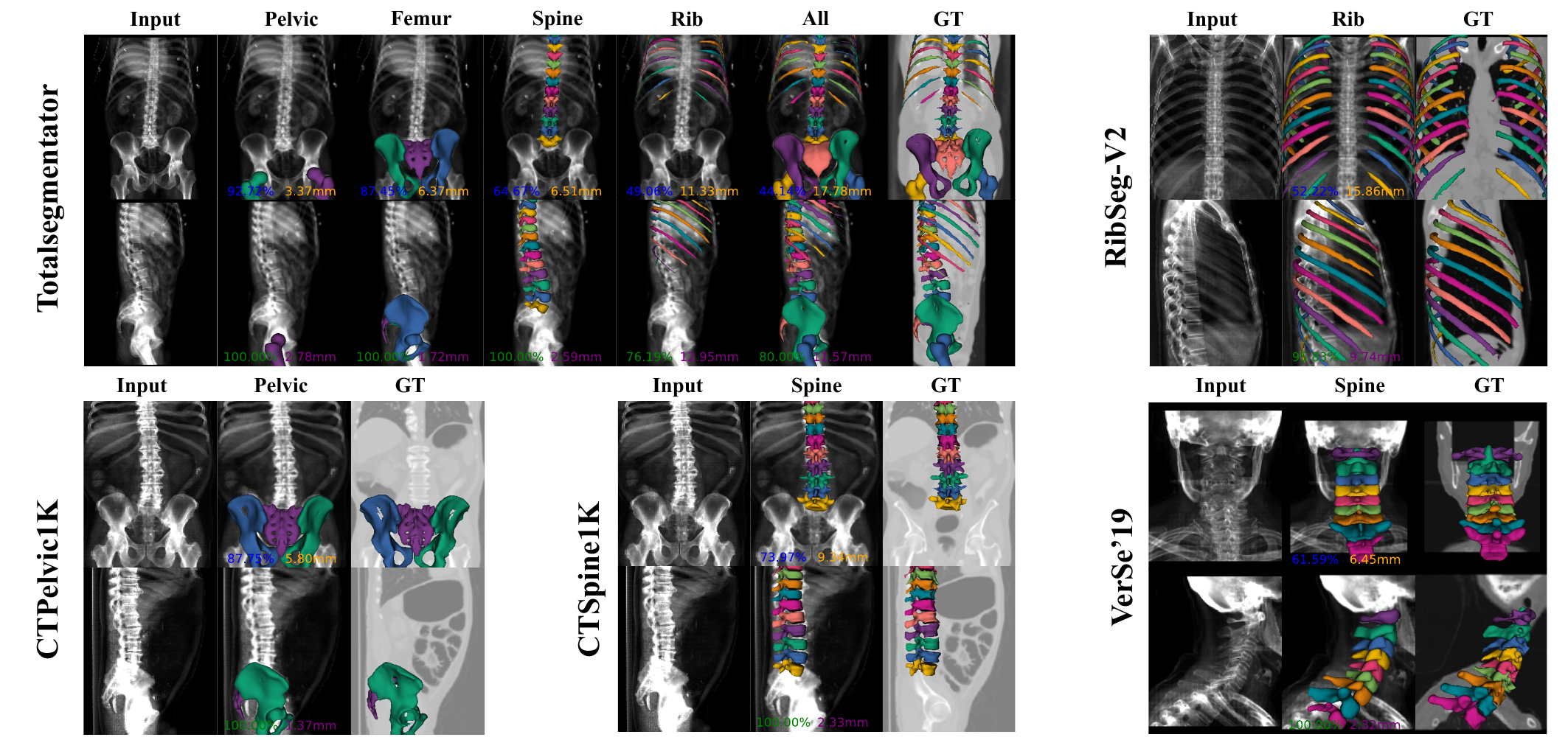}
	\caption{\small Quantitative reconstruction results of proposed Swin-X2S-Base network with biplanar inputs. Top-left panel: results of Totalsegmentator five subsets (femur, pelvis, spine, rib and all) on a single test sample. The first and second rows respectively exhibit the coronal and sagittal view. The first and last column denote biplanar inputs and ground truth, the other columns illustrate the segmentation results for different anatomies. The \textcolor{blue}{blue}, \textcolor[rgb]{1,0.647,0}{orange}, \textcolor[rgb]{0.0,0.5,0.0}{green} and \textcolor[rgb]{0.427,0.1,0.741}{purple} numbers respectively represent \textcolor{blue}{Dice}, \textcolor[rgb]{1,0.647,0}{HD}, \textcolor[rgb]{0.0,0.5,0.0}{L-error} and \textcolor[rgb]{0.427,0.1,0.741}{ID-rate}. Top-right panel: result of RibSeg v2 dataset on a single test sample. Bottom-left panel: result of CTPelvic1K. Bottom-center panel: result of CTSpine1K. Bottom-right panel: result of VerSe'19.}
	\label{fig03}
\end{figure*}

\subsection{Implementation details}
Due to the arbitrary field of view (FOV) in CT scans, which exhibit varying orientations, resolutions, spacings, and scanning modes, all samples were projected to synthesis biplanar X-ray images based on DRR generation method. Paired coronal and sagittal X-rays were resized to $128\times160$ pixels with equal spacing on each axis, ground truth CT segmentation were correspondingly resampled to isotropic $128\times128\times160$ voxels to ensure strict alignment with projections. Data augmentation strategies including random zoom, rotation, shift, and flipping were applied. 
Our algorithm was implemented with PyTorch and MONAI framework and trained on Nvidia Tesla A100 GPU with 40GB memory. Pre-trained Swin Transformer networks on ImageNet-22K was employed as Swin-X2S encoder backbone. The network batch size was set to 1 and the initial learning rate is set as 3e-5 with 5e-1 weight decay. We trained our model with the AdamW \cite{Loshchilov2018} optimizer for 250 epochs, incorporating with warm-up cosine scheduler for the first 20 epochs. 

\begin{figure}[ht]
	\centering
	\includegraphics[width=0.5\textwidth]{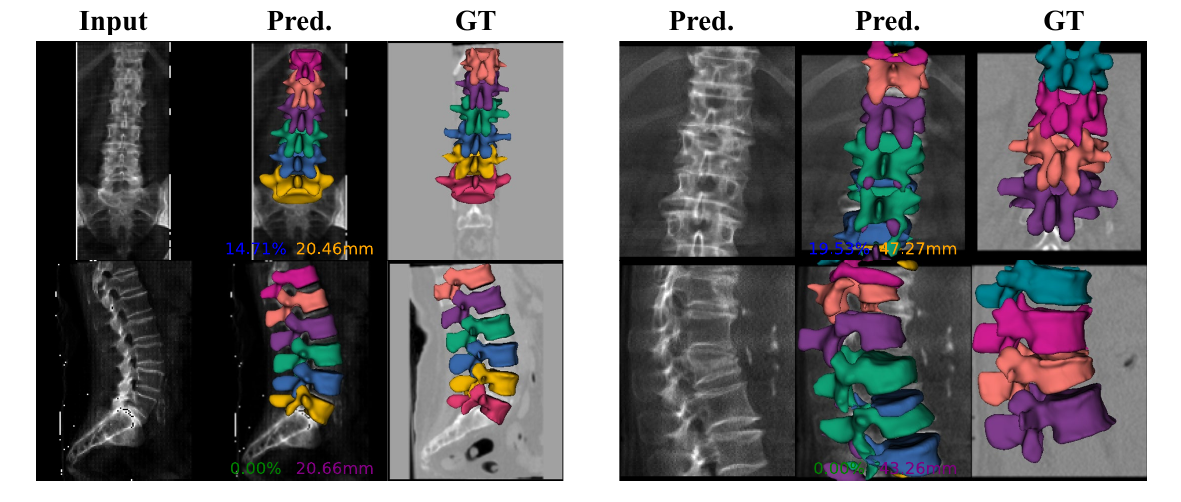}
	\caption{\small Reconstruction failure samples of Swin-X2S network on VerSe'19. The first, middle and last column denote biplanar inputs, prediction results and ground truth respectively}
	\label{fig04}
\end{figure}

\begin{figure*}[!t]
	\centering
	\includegraphics[width=1\textwidth]{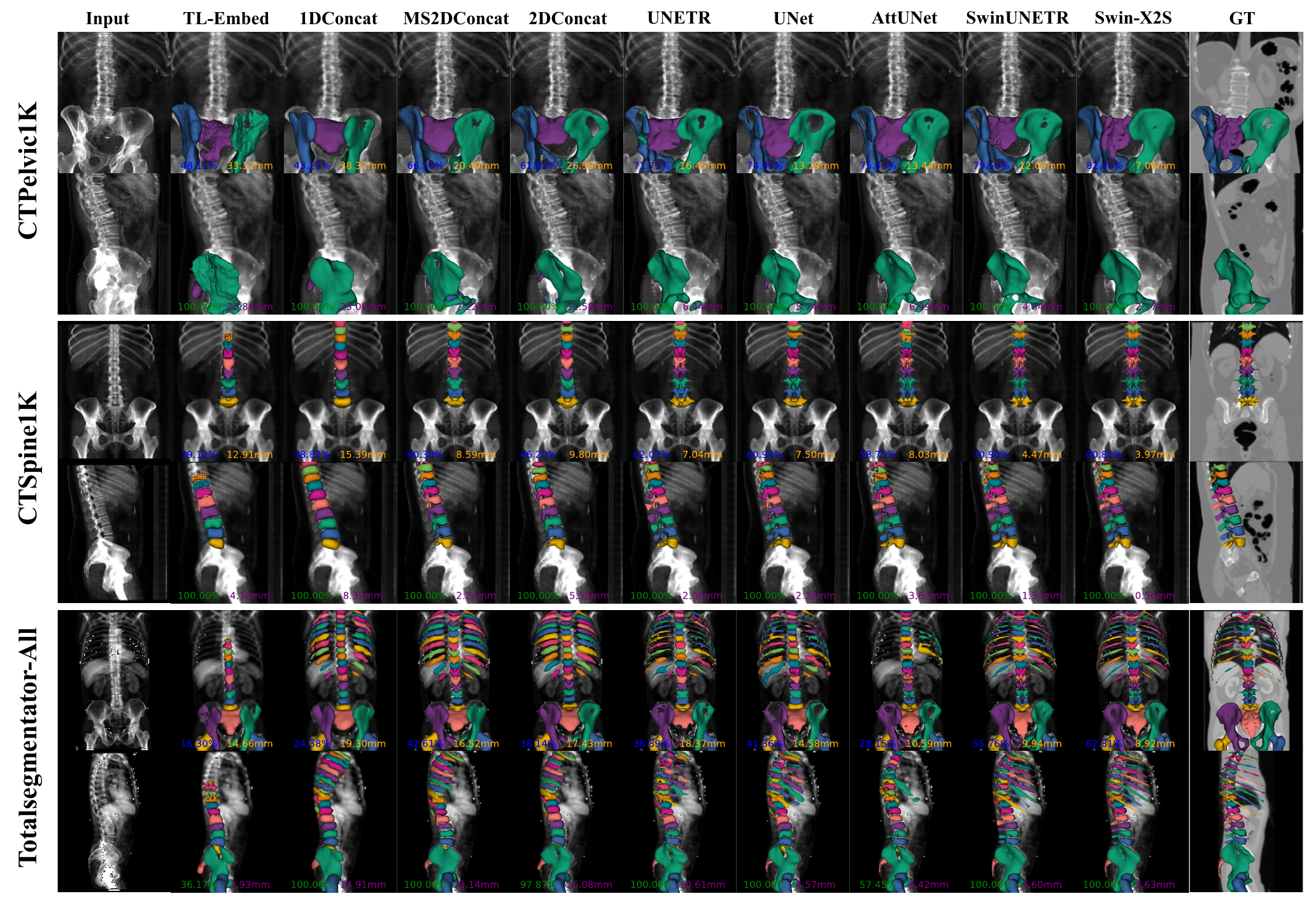}
	\caption{\small Comparison results of different methods on the CTPelvic1K, CTSpine1K, Totalsegmentatior-All datasets with biplanar inputs. The first and the last columns of each dataset denote biplanar inputs and ground truth, the other columns illustrate prediction results of different methods.}
	\label{fig05}
\end{figure*}

\begin{figure*}[!pt]
	\centering
	\includegraphics[width=1\textwidth]{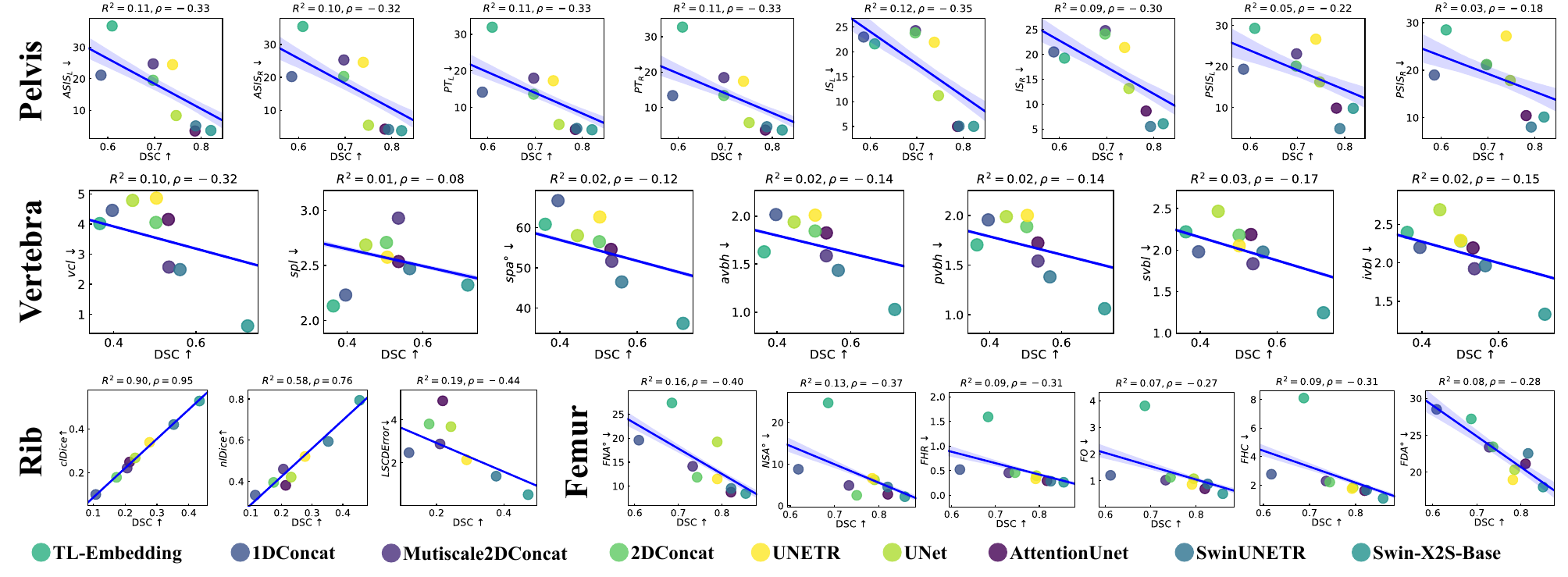}
	\caption{\small The correlation between the average Dice coefficient (x-axis) and clinical metrics (y-axis) on the Totalsegmentator-All dataset. The line denotes linear regression with confidence intervals of all reconstructed samples, and the scatter points represent the average value of corresponding algorithm's reconstructed samples.}
	\label{fig06}
\end{figure*}

\section{Result and analysis}
\subsection{Result on different datasets}
As shown in Table \ref{table02}, we evaluated the reconstruction performance across 9 subsets using Swin-X2S-Base network, the labeling results were directly derived from the segmentation masks. Despite the data patterns exhibiting diverse orientations, resolutions, spacings, and scanning modes, Swin-X2S demonstrates robust reconstruction performance on different structures, as illustrated in Fig. \ref{fig03}. It is noteworthy that the task of spine and rib reconstruction is particularly challenging due to their high structural similarities. Among the three subsets for spine segmentation and labeling, Swin-X2S achieved a mean Dice of 83.84\% on the CTSpine1K dataset, a mean Dice of 75.21\% on the Totalsegmentator-Spine dataset and a slightly lower mean Dice of 64.54\% for VerSe'19 dataset. The notable drop on the VerSe'19 dataset can be attributed to its relatively small data size, additionally, some samples were excessively cropped led to the absence contextual information, which in turn resulted the difficulties on vertebra identification (Fig. \ref{fig04}). Due to the inherently slender structure of ribs making them even harder to localize, proposed model achieves the lowest Dice of 45.92\% and 56.43\% on RibSeg v2 and Totalsegmentator-Rib dataset among all four anatomies. Most samples of Totalsegmentator-Rib exhibit modality diversity which resulting in incomplete rib cages and posing additional challenges for identification.

Moreover, we directly reconstruct all above four anatomies in the Totalsegmentator-All dataset. Proposed method achieves a mean Dice of 57.51\% and mean HD 13.33 mm for segmentation, a mean L-error of 9.22 mm and mean ID-rate 83.56\% for labeling. Overall, the performance of modeling all anatomies together is slightly lower than modeling each structure separately (Fig. \ref{fig06}).

\subsection{Result on different methods}
As shown in Table \ref{table03}, we evaluated the reconstruction performance of several existing methods on CTPelvic1K, CTSpine1K, and Totalsegmentator-All dataset. The comparison methods adhered to the settings outlined in \cite{Shakya2024}. Notably, the regression head used for generating binary maps was replaced with classification head for segmentation and labeling. 

Across these three tasks, Swin-X2S models achieve the state of art reconstruction performance. Previous deep learning methods suffer from the challenge of category confusion and fuzzy boundaries, where adjacent structures are not precisely separated by two boundary interfaces, as shown in Fig. \ref{fig05}, leading to segmentation inconsistencies inside anatomy. Transformer-based architecture, such as SwinUNETR \cite{Hatamizadeh2021} and AttentionUNet \cite{Oktay2018} generally performed better than pure CNN-based architecture. This superiority can be attributed to self-attention mechanism which captures global context effectively, while convolution mechanism inherently focuses on local spatial features. Feature-fusion methods like 2DConcat \cite{Bayat2020}, UNet \cite{Hatamizadeh2022} and Mutiscale2DConcat \cite{Ying2019} preformed better than embedding vector fusion methods like 1DConcat \cite{Chen2020} and TL-Embedding \cite{Girdhar2016}. Fig. \ref{fig05} visualizes comparison results between the ground truth segmentation and the shape reconstruction achieved by different networks. It can be seen that even for shifted, rotated, scaled and deformed structures, our method is able to segment category boundaries clearly, achieve the best reconstruction results among all algorithms.

Moreover, we present quantitative results of Swin-X2S models with various architectural settings across these datasets. Swin-X2S-Tiny achieved the lowest mean Dice across these three reconstruction tasks. In comparison, Swin-X2S-Base performs slightly better than Swin-X2S-Small, with more substantial performance improvements observed for more difficult tasks. Despite the model size of Swin-X2S-Large is twice that of Swin-X2S-Base, its performance has barely improved. These experiments demonstrate the success of Swin-X2S owes not only to the larger number of parameters but also to the appropriate architecture settings.

\begin{table*}[t]
	\fontsize{8}{10}\selectfont
	\centering
	\caption{\small Comparison results of different models on the CTPelvic1K, CTSpine1K and Totalsegmentator-All datasets using biplanar inputs. The \textbf{bolded} numbers denote the highest score and the \textit{italicized} numbers indicate the second highest.}
	\label{table03}
	\setlength{\tabcolsep}{1.5pt}
	\renewcommand\arraystretch{1.2}
	\begin{tabular}{l|cccc|cccc|cccc}
		\hline
		\multirow{2}{*}{Model (reference)} & \multicolumn{4}{c|}{CTPelvic1K-dataset} & \multicolumn{4}{c|}{CTSpine1K-dataset} & \multicolumn{4}{c}{Totalsegmentator-All-dataset} \\ \cline{2-13} 
		& \multicolumn{1}{l}{Dice(\%)} & \multicolumn{1}{l}{HD(mm)} & \multicolumn{1}{l}{L-error(mm)} & \multicolumn{1}{l|}{ID-rate(\%)} & \multicolumn{1}{l}{Dice(\%)} & \multicolumn{1}{l}{HD(mm)} & \multicolumn{1}{l}{L-error(mm)} & \multicolumn{1}{l|}{ID-rate(\%)} & \multicolumn{1}{l}{Dice(\%)} & \multicolumn{1}{l}{HD(mm)} & \multicolumn{1}{l}{L-error(mm)} & \multicolumn{1}{l}{ID-rate(\%)} \\ \hline
		TL-Embedding \cite{Girdhar2016} & 69.02 & 28.84 & 8.92 & 95.63 & 44.33 & 17.51 & 11.10 & 82.15 & 18.56 & 23.31 & - & 69.98 \\
		1DConcat \cite{Chen2020} & 69.57 & 20.39 & 9.75 & 93.51 & 53.64 & 17.38 & 10.45 & 81.51 & 27.41 & 41.51 & 25.23 & 53.65 \\ 
		Mutiscale2DConcat \cite{Ying2019} & 81.06 & 8.92 & 1.78 & 98.79 & 72.74 & 10.40 & 5.00 & 81.94 & 39.14 & 34.69 & 20.82 & 60.27 \\
		2DConcat \cite{Bayat2020} & 80.77 & 9.51 & 5.02 & 98.79 & 63.35 & 12.88 & 6.69 & 92.34 & 35.72 & 64.79 & 23.72 & 57.93 \\
		UNETR \cite{Hatamizadeh2022} & 83.42 & 9.41 & 4.44 & 99.10 & 63.51 & 16.53 & 8.56 & 71.99 & 37.84 & 37.18 & 20.78 & 58.79 \\
		UNet \cite{Kasten2020} & 84.26 & 7.88 & 3.94 & 98.79 & 69.15 & 11.77 & 5.39 & 73.94 & 42.19 & 21.07 & 14.44 & 73.35 \\
		AttentionUnet \cite{Oktay2018} & 84.90 & 7.45 & 3.84 & 98.64 & 69.53 & 10.36 & 5.21 & 81.47 & 33.08 & 19.03 & 10.15 & 82.84 \\
		SwinUNETR \cite{Hatamizadeh2021} & 87.75 & 5.83 & 2.95 & 98.94 & 77.13 & 8.47 & 4.10 & 79.26 & 48.66 & 19.57 & 13.15 & 73.16 \\ \hline
		Swin-X2S-Tiny & 83.20 & 5.31 & 2.95 & 98.52 & 74.44 & 8.47 & 5.10 & 82.58 & 44.81 & 22.46 & 15.31 & 79.57 \\
		Swin-X2S-Small & 88.72 & 4.52 & 2.78 & 97.21 & 80.38 & 6.44 & 3.69 & \textbf{85.67} & 52.87 & 16.81 & 12.58 & 82.11 \\
		Swin-X2S-Base & \textit{89.11} &  \textit{4.66} & \textbf{2.57} & \textit{98.79} & \textbf{83.84} & \textbf{5.27} & \textbf{2.79} & 84.76 & \textit{57.51} & \textit{13.33} & \textit{9.22} & \textbf{83.56} \\
		Swin-X2S-Large & \textbf{89.20} & \textbf{4.41} & \textit{2.60} & \textbf{99.21} & \textit{83.41} & \textit{5.44} & \textit{2.97} & \textit{85.18} & \textbf{57.83} & \textbf{12.84} & \textbf{9.10} & \textit{82.97} \\ \hline
	\end{tabular}
\end{table*}

\subsection{Result on clinical metrics}
Although automated shape feature extraction pipelines are robust at large, they still encounter additional errors introduced by mesh distortion and inaccurate localization. We evaluate the clinical metrics on Totalsegmentator-All as shown in Fig. \ref{fig06}. The quality of clinical shape estimation is generally positively correlated with the average Dice, however, there are exceptions such as \textbf{(i)} proposed Swin-X2S achieves the highest Dice on all four anatomies but performing worse than SwinUNETR in pelvic landmark especially for posterior superior iliac spine (PSIS); \textbf{(ii)} TL-Embedding and 1DConcat possess similar mean Dice, TL-Embedding completely fails in rib cage reconstruction, while 1DConcat performs much better in evaluating clinical parameters and almost achieving results comparable to 2DConcat on pelvis and femur morphometry.

Swin-X2S achieves the best femoral head radius (FHR) and neck shaft angle (NSA) evaluation, which are crucial for the diagnosis of hip deformities and guidance of surgical planning for hip replacement.
Pelvic landmarks, which are muscle and ligament attachment points, serve as key references for imaging diagnosis. SwinUNETR and AttentionUnet achieve similar Anterior Superior Iliac Spine (ASIS) localization compare to proposed method, even performs better on posterior ilium region. 
The correlation between vertebra parameters and Dice is the lowest, this is because we reconstruct the whole spine and then divide it into individual vertebrae, resulting in pixelated shape that introduces further challenges for evaluation. Nevertheless, Swin-X2S still shows significant improvements in determining the pedicle of vertebral arch area (vertebral canal length, vcl), which is the safe zone for pedicle screw insertion in spinal surgery.
Although the average Dice of the ribs is the lowest among all structures, we observed that Swin-X2S achieved over 80\% in nlDice, indicating that the main factor affecting shape accuracy is the axial deviation of centerline (LSCDError) rather than the curvature deviation of morphology (nlDice). The latter is a more valuable indicator for fracture detection and spinal deformities.

\begin{figure*}[!t]
	\centering
	\includegraphics[width=1\textwidth]{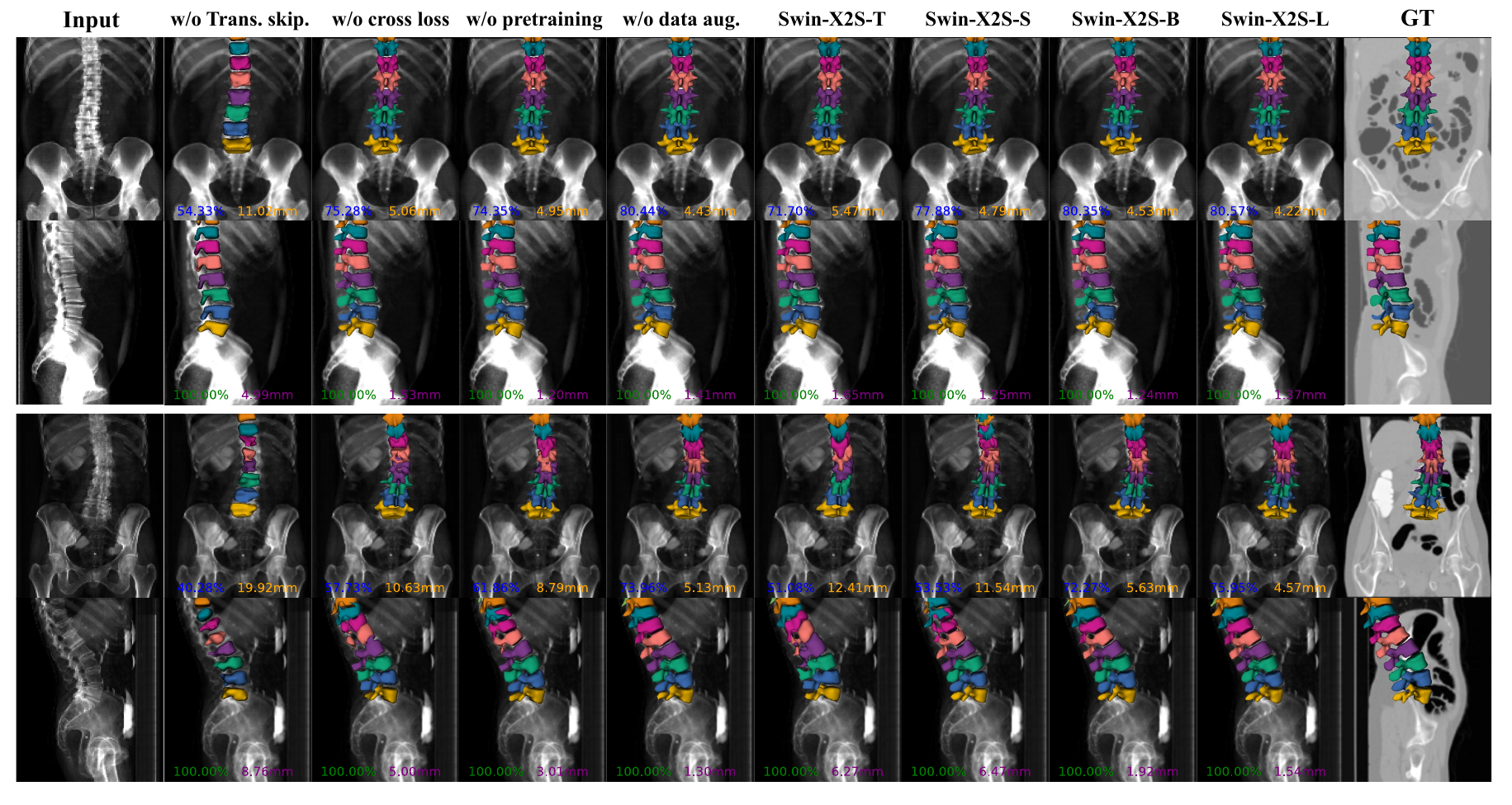}
	\caption{Ablation study of the key components of Swin-X2S on the CTSpine1K dataset with biplanar DRR inputs. Visualization results do not include the untrainable Swin-X2S-Base model without convolution skip connections.}
	\label{fig07}
\end{figure*}

\begin{table}[!b]
	\fontsize{8}{10}\selectfont
	\centering
	\setlength{\tabcolsep}{3pt}
	\renewcommand\arraystretch{1.2}
	\caption{\small Ablation study on the influence of the number of DRR views $N$ on CTSpine1K dataset using Swin-X2S-Base network.}
	\label{table04}
	\begin{tabular}{c|cccc}
		\hline
		\multicolumn{1}{l|}{\multirow{1}{*}{Num of views}} & \multirow{1}{*}{Dice(\%)↑} & \multirow{1}{*}{HD(mm)↓} & \multirow{1}{*}{L-error(mm)↓} & \multirow{1}{*}{ID-rate(\%)↑} \\ \hline
		\multirow{1}{*}{$N=1$} & 83.84 & 5.27 & \textbf{2.79} & \textit{87.76} \\ \hline
		\multirow{1}{*}{$N=2$} & 83.23 & 5.28 & 2.97 & 87.19 \\ \hline
		\multirow{1}{*}{$N=4$} & 83.99 & 5.20 & \textit{2.89} & \textbf{87.67} \\ \hline
		\multirow{1}{*}{$N=10$}	& \textbf{84.04} & \textit{5.05} & 3.08 & 87.58 \\ \hline
		\multirow{1}{*}{$N=20$} & \textit{83.81} & \textbf{4.95} & 3.00 & 87.33 \\ \hline
	\end{tabular}
\end{table}
\subsection{Ablation study}
\subsubsection{The number of DRR views}
We investigated the impact of the number of DRR views on performance in Table \ref{table04}. As the number of projected views increases, there is a slight improvement (less than 1\% improvement on Dice) in segmentation and annotation performance. This improvement can be attributed to the additional contextual information provided by the multi-view projections, which help to supplement occluded areas. However, non-coronal or non-sagittal images are unable aligned with ground truth masks, which may explain the limited marginal performance enhancement. Furthermore, more views come at the cost of proportional radiation dose and scanning time. Therefore, biplanar inputs are preferred for 2D-3D reconstruction task.

\subsubsection{The effectiveness of key components}
We perform ablation study ti access the impact of key components of Swin-X2S on CTSpine1K using biplanar inputs (Table \ref{table05}). First, we examined the effect of skip connections, the absence of transformer skip connections resulted in a significantly performance drop, with the mean Dice drop from 83.84\% to 61.79\%. The absence of convolution skip connections caused severe issues, leading to an untrainable model due to mode collapse. Unsupervised cross loss also plays an important role in training process, sole reliance on supervised loss lead to a noticeable mean Dice drop of 5.76\%. Additionally, pre-trained Swin Transformer weights on ImageNet-22K facilitate network's ability to learn generalized feature representations, resulting in improved reconstruction performance and faster convergence during training process, the base model experienced a drop 4.44\% of mean Dice without pre-trained weights. Data augmentation strategies are crucial for generalization ability, particularly for conditions like scoliosis, fracture and mental implants. The absence of this strategy lead to a decrease of 1.04\% in segmentation (Dice) and an increase of 0.28 mm in labeling (L-error).

\begin{table}[!h]
	\fontsize{8}{10}\selectfont
	\centering
	\setlength{\tabcolsep}{3pt}
	\renewcommand\arraystretch{1.2}
	\caption{\small Ablation study on the effect of model key components on CTSpine1K dataset using biplanar inputs. Trans. and conv. skip. mean transformer and convolution skip connection.}
	\label{table05}
	\begin{tabular}{l|cc}
		\hline
		\multirow{1}{*}{Model architecture} & \multirow{1}{*}{\qquad Dice(\%)} & \multirow{1}{*}{\qquad L-error(mm) \qquad}  \\ \hline
		Swin-X2S-Base w/o Trans. skip. & 61.79 & 8.49 \\
		Swin-X2S-Base w/o conv. skip.  & -  & - \\
		Swin-X2S-Base w/o cross loss  & 78.08 &  4.12 \\
		Swin-X2S-Base w/o pre-training  & 79.30  &  3.29 \\ 
		Swin-X2S-Base w/o data aug.  & 82.80 & 3.07 \\ \hline
		Swin-X2S-Tiny & 74.44  & 5.10  \\ 
		Swin-X2S-Small & 80.88  & 3.66  \\ 
		Swin-X2S-Base & \textbf{83.84}  & \textbf{2.79}  \\ 
		Swin-X2S-Large & \textit{83.41}  & \textit{2.97}  \\ \hline
	\end{tabular}
\end{table}

\section{Discussion}
We introduced a novel end to end method for the reconstruction of 3D bone shapes based on 2D X-ray images. Thoroughly experiments were evaluated across nine publicly available datasets covering four categories of skeleton anatomies. Experimental results showed that our method achieved the state of art reconstruction performance in both segmentation and labeling tasks. In terms of clinical metrics, Swin-X2S demonstrates overall satisfactory performance, however, we observed that shape features are influenced by relatively unstable morphometry extraction pipelines, highlighting the need for further development of robust clinical parameter estimation methods to ensure effective evaluation in real-world scenarios.

DRR generation serves as a compromise solution due to the lack of datasets consisting of paired real X-ray images and aligned CT scans. The collection of such a dataset becomes an urgent task for reconstruction method research, requiring attention and effort. A collaborative effort from the community could help make these datasets more accessible to the public.

2D-3D reconstruction not only can be applied in preliminary diagnosis and treatment, it also holds potential for real-time 3D visualization in orthopedic surgeries assisted by mobile C-arm digital radiography in the future. Furthermore, our current work focuses solely on bone shape reconstruction, which could also be used to quickly locate and reshape foreign objects, such as metal implants or fragments. Exploring the reconstruction methods for non-osseous anatomical structures may also hold significant value of research.
\newline
\newline
\newline

\section*{Appendix}
\section*{Appendix A. Comparison Network Architecture} \label{sec:method}
All seven comparison model architectures were illustrated in table \ref{table06} , the image resolution is set to [128, 128, 160] except for TL-Embedding and 1DConcat due to architecture constraint, the batch size was set to 1 for SwinUNETR and UNETR, 4 for MultiScale2DConcat and 8 for others, the initial learning rate is set as 2e-3 for UNETR and SwinUNETR and 2e-4 for the other models. For each dataset, the classification head was configured to corresponding class number, all architectures were trained with the AdamW optimizer for 50 epochs, with a warm-up cosine scheduler for the first 5 epochs.

\begin{table}[!h]
	\fontsize{8}{10}\selectfont
	\centering
	\setlength{\tabcolsep}{3pt}
	\renewcommand\arraystretch{1.2}
	\caption{The model hyperparameters of the comparison methods.}
	\label{table06}
	\begin{tabular}{ll|ccc}
		\hline
		\multicolumn{2}{l|}{\multirow{2}{*}{Model Architecture \quad}} & \multicolumn{1}{l}{\multirow{2}{*}{Image Size}} & \multirow{2}{*}{Paramethers(M)} & \multirow{2}{*}{FLOPs(G)} \\
		\multicolumn{2}{l|}{} & \multicolumn{1}{l}{} &  &  \\ \hline
		\multicolumn{2}{l|}{TL-Embedding} & {[}128,128,128{]} & 53.65 & 4.38 \\ \hline
		\multicolumn{2}{l|}{1DConcat} & {[}128,128,128{]} & 40.67 & 256.08 \\ \hline
		\multicolumn{2}{l|}{Multiscale2DConcat} & {[}128,128,160{]} & 3.08 & 365.55 \\ \hline
		\multicolumn{2}{l|}{2DConcat} & {[}128,128,160{]} & 1.55 & 1737.96 \\ \hline
		\multicolumn{2}{l|}{UNETR} & {[}128,128,160{]} & 95.77 & 499.92 \\ \hline
		\multicolumn{2}{l|}{UNet} & {[}128,128,160{]} & 1.31 & 569.06 \\ \hline
		\multicolumn{2}{l|}{AttentionNet} & {[}128,128,160{]} & 1.48 & 96.06 \\ \hline
		\multicolumn{2}{l|}{SwinUNETR} & {[}128,128,160{]} & 61.99 & 1978.90 \\ \hline	
	\end{tabular}
\end{table}

\begin{figure}[!t]
	\centering
	\includegraphics[width=0.5\textwidth]{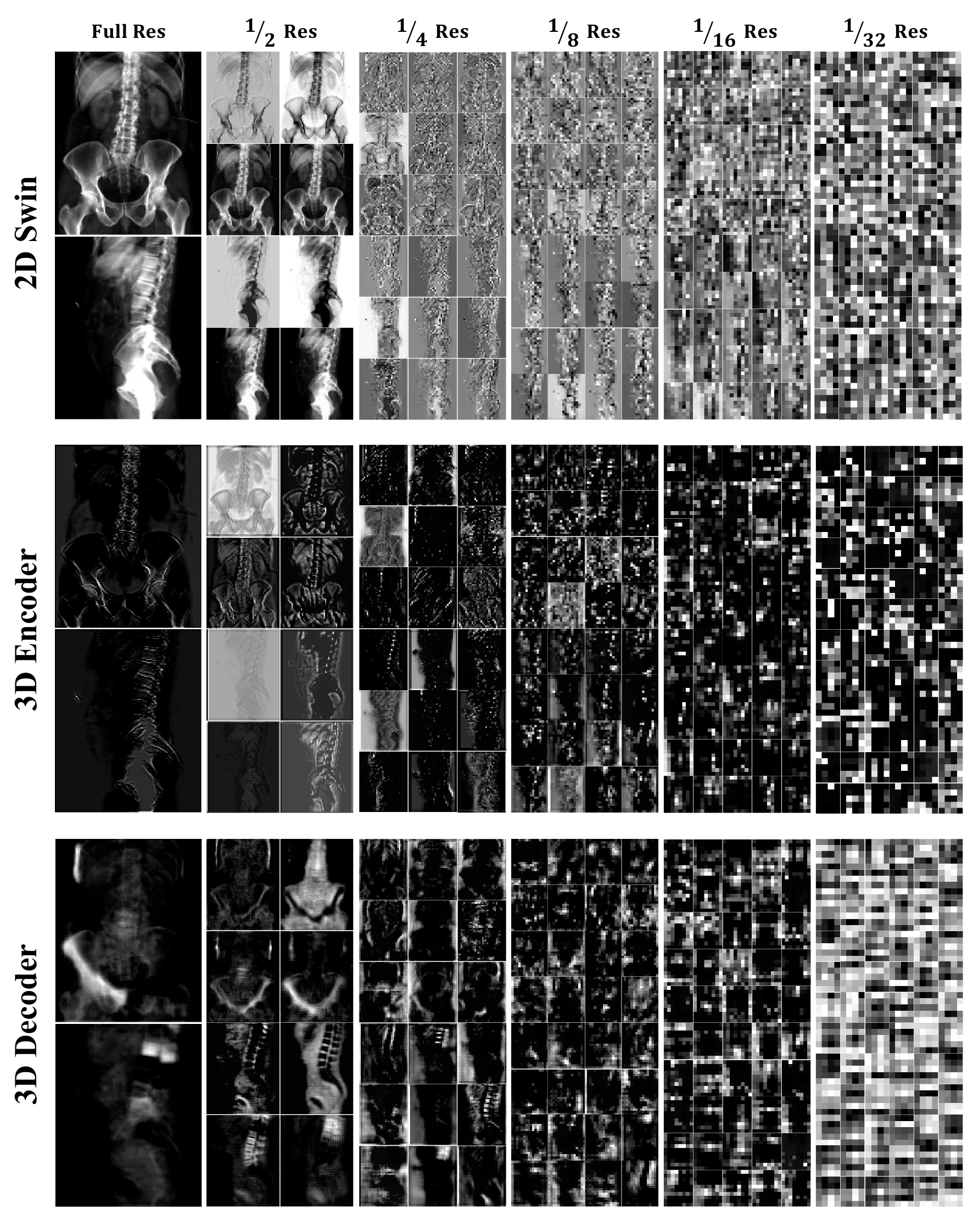}
	\caption{\small Visualization results of feature maps in Swin-X2S-Base skip connections. The first row, the middle row, and the last row denote the output of 2D Swin Transformer, 3D encoder, and 3D decoder at each stage, respectively.}
	\label{fig08}
\end{figure}

\section*{Appendix B. Visualization of Feature Maps}
As shown in fig. \ref{fig08}, we present the visualization results of feature maps in the Swin-X2S-Base skip connections trained on Totalsegmentator-All dataset. Swin-X2S consists of six stages, with each subsequent stage having half the resolution and twice the number of channels of the previous stage. The rows perspective exhibits that the deeper features possess the more compressed and abstract forms of expression, the columns perspective shows 2D Swin Transformer extract the basic features of X-rays, while 3D encoder and decoder focus on the segmentation and labeling of interest anatomies.

\normalsize
\bibliography{references}

\begin{thebibliography}{54}
\providecommand{\natexlab}[1]{#1}
\providecommand{\url}[1]{\texttt{#1}}
\expandafter\ifx\csname urlstyle\endcsname\relax
  \providecommand{\doi}[1]{doi: #1}\else
  \providecommand{\doi}{doi: \begingroup \urlstyle{rm}\Url}\fi

\bibitem[Brown et~al.(1976)Brown, Burstein, Nash, and Schock]{Brown1976}
Richard~H Brown, Albert~H Burstein, Clyde~L Nash, and Charles~C Schock.
\newblock Spinal analysis using a three-dimensional radiographic technique.
\newblock \emph{Journal of biomechanics}, 9\penalty0 (6):\penalty0 355--IN1,
  1976.

\bibitem[Herman(2009)]{Herman2009}
Gabor~T Herman.
\newblock \emph{Fundamentals of computerized tomography: image reconstruction
  from projections}.
\newblock Springer Science \& Business Media, 2009.

\bibitem[Hosseinian and Arefi(2015)]{Hosseinian2015}
S~Hosseinian and H~Arefi.
\newblock 3d reconstruction from multi-view medical x-ray images--review and
  evaluation of existing methods.
\newblock \emph{The international archives of the photogrammetry, remote
  sensing and spatial information sciences}, 40:\penalty0 319--326, 2015.

\bibitem[Ying et~al.(2019)Ying, Guo, Ma, Wu, Weng, and Zheng]{Ying2019}
Xingde Ying, Heng Guo, Kai Ma, Jian Wu, Zhengxin Weng, and Yefeng Zheng.
\newblock X2ct-gan: reconstructing ct from biplanar x-rays with generative
  adversarial networks.
\newblock In \emph{Proceedings of the IEEE/CVF conference on computer vision
  and pattern recognition}, pages 10619--10628, 2019.

\bibitem[Shakya and Khanal(2024)]{Shakya2024}
Mahesh Shakya and Bishesh Khanal.
\newblock Benchmarking encoder-decoder architectures for biplanar x-ray to 3d
  bone shape reconstruction.
\newblock \emph{Advances in Neural Information Processing Systems}, 36, 2024.

\bibitem[Hricak et~al.(2021)Hricak, Abdel-Wahab, Atun, Lette, Paez, Brink,
  Donoso-Bach, Frija, Hierath, Holmberg, et~al.]{Hricak2021}
Hedvig Hricak, May Abdel-Wahab, Rifat Atun, Miriam~Mikhail Lette, Diana Paez,
  James~A Brink, Llu{\'\i}s Donoso-Bach, Guy Frija, Monika Hierath, Ola
  Holmberg, et~al.
\newblock Medical imaging and nuclear medicine: a lancet oncology commission.
\newblock \emph{The Lancet Oncology}, 22\penalty0 (4):\penalty0 e136--e172,
  2021.

\bibitem[Ngoya et~al.(2016)Ngoya, Muhogora, and Pitcher]{Ngoya2016}
Patrick~Sitati Ngoya, Wilbroad~Edward Muhogora, and Richard~Denys Pitcher.
\newblock Defining the diagnostic divide: an analysis of registered
  radiological equipment resources in a low-income african country.
\newblock \emph{The Pan African Medical Journal}, 25, 2016.

\bibitem[Aubert et~al.(2019)Aubert, Vazquez, Cresson, Parent, and
  de~Guise]{Aubert2019}
Benjamin Aubert, Carlos Vazquez, Thierry Cresson, Stefan Parent, and Jacques~A
  de~Guise.
\newblock Toward automated 3d spine reconstruction from biplanar radiographs
  using cnn for statistical spine model fitting.
\newblock \emph{IEEE transactions on medical imaging}, 38\penalty0
  (12):\penalty0 2796--2806, 2019.

\bibitem[Ill{\'e}s and Somoske{\"o}y(2012)]{Illes2012}
Tam{\'a}s Ill{\'e}s and Szabolcs Somoske{\"o}y.
\newblock The eos™ imaging system and its uses in daily orthopaedic practice.
\newblock \emph{International orthopaedics}, 36:\penalty0 1325--1331, 2012.

\bibitem[Krizhevsky et~al.(2012)Krizhevsky, Sutskever, and
  Hinton]{Krizhevsky2012}
Alex Krizhevsky, Ilya Sutskever, and Geoffrey~E Hinton.
\newblock Imagenet classification with deep convolutional neural networks.
\newblock \emph{Advances in neural information processing systems}, 25, 2012.

\bibitem[He et~al.(2016)He, Zhang, Ren, and Sun]{He2016}
Kaiming He, Xiangyu Zhang, Shaoqing Ren, and Jian Sun.
\newblock Deep residual learning for image recognition.
\newblock In \emph{Proceedings of the IEEE conference on computer vision and
  pattern recognition}, pages 770--778, 2016.

\bibitem[Vaswani et~al.(2017)Vaswani, Shazeer, Parmar, Uszkoreit, Jones, Gomez,
  Kaiser, and Polosukhin]{Vaswani2017}
Ashish Vaswani, Noam Shazeer, Niki Parmar, Jakob Uszkoreit, Llion Jones,
  Aidan~N Gomez, {\L}ukasz Kaiser, and Illia Polosukhin.
\newblock Attention is all you need.
\newblock \emph{Advances in neural information processing systems}, 30, 2017.

\bibitem[Maken and Gupta(2023)]{Maken2023}
P~Maken and A~Gupta.
\newblock 2d-to-3d: A review for computational 3d image reconstruction from
  x-ray images.
\newblock \emph{DOI: https://doi. org/10.1007/s11831-022-09790-z}, pages
  85--114, 2023.

\bibitem[Mitton et~al.(2000)Mitton, Landry, Veron, Skalli, Lavaste, and
  De~Guise]{Mitton2000}
D~Mitton, C~Landry, S~Veron, Wata Skalli, F~Lavaste, and Jacques~A De~Guise.
\newblock 3d reconstruction method from biplanar radiography using
  non-stereocorresponding points and elastic deformable meshes.
\newblock \emph{Medical and Biological Engineering and Computing}, 38:\penalty0
  133--139, 2000.

\bibitem[Mitulescu et~al.(2002)Mitulescu, Skalli, Mitton, and
  De~Guise]{Mitulescu2002}
A~Mitulescu, Wata Skalli, D~Mitton, and J~De~Guise.
\newblock Three-dimensional surface rendering reconstruction of scoliotic
  vertebrae using a non stereo-corresponding points technique.
\newblock \emph{European spine journal}, 11:\penalty0 344--352, 2002.

\bibitem[Benameur et~al.(2003)Benameur, Mignotte, Parent, Labelle, Skalli, and
  de~Guise]{Benameur2003}
Said Benameur, Max Mignotte, Stefan Parent, Hubert Labelle, Wafa Skalli, and
  Jacques de~Guise.
\newblock 3d/2d registration and segmentation of scoliotic vertebrae using
  statistical models.
\newblock \emph{Computerized Medical Imaging and Graphics}, 27\penalty0
  (5):\penalty0 321--337, 2003.

\bibitem[Boisvert and Moura(2011)]{Boisvert2011}
Jonathan Boisvert and Daniel~C. Moura.
\newblock Interactive 3d reconstruction of the spine from radiographs using a
  statistical shape model and second-order cone programming.
\newblock In \emph{2011 Annual International Conference of the IEEE Engineering
  in Medicine and Biology Society}, pages 5726--5729, 2011.
\newblock \doi{10.1109/IEMBS.2011.6091386}.

\bibitem[Whitmarsh et~al.(2013)Whitmarsh, Humbert, Barquero, Di~Gregorio, and
  Frangi]{Whitmarsh2013}
Tristan Whitmarsh, Ludovic Humbert, Luis M Del~R{\'\i}o Barquero, Silvana
  Di~Gregorio, and Alejandro~F Frangi.
\newblock 3d reconstruction of the lumbar vertebrae from anteroposterior and
  lateral dual-energy x-ray absorptiometry.
\newblock \emph{Medical image analysis}, 17\penalty0 (4):\penalty0 475--487,
  2013.

\bibitem[Karade and Ravi(2015)]{Karade2015}
Vikas Karade and Bhallamudi Ravi.
\newblock 3d femur model reconstruction from biplane x-ray images: a novel
  method based on laplacian surface deformation.
\newblock \emph{International journal of computer assisted radiology and
  surgery}, 10:\penalty0 473--485, 2015.

\bibitem[Anas et~al.(2016)Anas, Rasoulian, Seitel, Darras, Wilson, John,
  Pichora, Mousavi, Rohling, and Abolmaesumi]{Anas2016}
Emran Mohammad~Abu Anas, Abtin Rasoulian, Alexander Seitel, Kathryn Darras,
  David Wilson, Paul~St John, David Pichora, Parvin Mousavi, Robert Rohling,
  and Purang Abolmaesumi.
\newblock Automatic segmentation of wrist bones in ct using a statistical wrist
  shape $+ $ pose model.
\newblock \emph{IEEE transactions on medical imaging}, 35\penalty0
  (8):\penalty0 1789--1801, 2016.

\bibitem[Ch{\^e}nes and Schmid(2021)]{Chenes2021}
Christophe Ch{\^e}nes and J{\'e}r{\^o}me Schmid.
\newblock Revisiting contour-driven and knowledge-based deformable models:
  Application to 2d-3d proximal femur reconstruction from x-ray images.
\newblock In \emph{Medical Image Computing and Computer Assisted
  Intervention--MICCAI 2021: 24th International Conference, Strasbourg, France,
  September 27--October 1, 2021, Proceedings, Part VI 24}, pages 451--460,
  2021.

\bibitem[Song et~al.(2021)Song, Liang, Yang, Wang, and He]{Song2021}
Weinan Song, Yuan Liang, Jiawei Yang, Kun Wang, and Lei He.
\newblock Oral-3d: Reconstructing the 3d structure of oral cavity from
  panoramic x-ray.
\newblock In \emph{Proceedings of the AAAI conference on artificial
  intelligence}, volume~35, pages 566--573, 2021.

\bibitem[Shen et~al.(2019)Shen, Zhao, and Xing]{Shen2019}
Liyue Shen, Wei Zhao, and Lei Xing.
\newblock Patient-specific reconstruction of volumetric computed tomography
  images from a single projection view via deep learning.
\newblock \emph{Nature biomedical engineering}, 3\penalty0 (11):\penalty0
  880--888, 2019.

\bibitem[Ge et~al.(2022)Ge, He, Xia, Xu, Sun, Yang, Li, Wang, Yu, Zhang,
  et~al.]{Ge2022}
Rongjun Ge, Yuting He, Cong Xia, Chenchu Xu, Weiya Sun, Guanyu Yang, Junru Li,
  Zhihua Wang, Hailing Yu, Daoqiang Zhang, et~al.
\newblock X-ctrsnet: 3d cervical vertebra ct reconstruction and segmentation
  directly from 2d x-ray images.
\newblock \emph{Knowledge-Based Systems}, 236:\penalty0 107680, 2022.

\bibitem[Kyung et~al.(2023)Kyung, Jo, Choo, Lee, and Choi]{Kyung2023}
Daeun Kyung, Kyungmin Jo, Jaegul Choo, Joonseok Lee, and Edward Choi.
\newblock Perspective projection-based 3d ct reconstruction from biplanar
  x-rays.
\newblock In \emph{ICASSP 2023-2023 IEEE International Conference on Acoustics,
  Speech and Signal Processing (ICASSP)}, pages 1--5, 2023.

\bibitem[Kasten et~al.(2020)Kasten, Doktofsky, and Kovler]{Kasten2020}
Yoni Kasten, Daniel Doktofsky, and Ilya Kovler.
\newblock End-to-end convolutional neural network for 3d reconstruction of knee
  bones from bi-planar x-ray images.
\newblock In \emph{Machine Learning for Medical Image Reconstruction: Third
  International Workshop, MLMIR 2020, Held in Conjunction with MICCAI 2020,
  Lima, Peru, October 8, 2020, Proceedings 3}, pages 123--133, 2020.

\bibitem[Shiode et~al.(2021)Shiode, Kabashima, Hiasa, Oka, Murase, Sato, and
  Otake]{Shiode2021}
Ryoya Shiode, Mototaka Kabashima, Yuta Hiasa, Kunihiro Oka, Tsuyoshi Murase,
  Yoshinobu Sato, and Yoshito Otake.
\newblock 2d--3d reconstruction of distal forearm bone from actual x-ray images
  of the wrist using convolutional neural networks.
\newblock \emph{Scientific Reports}, 11\penalty0 (1):\penalty0 15249, 2021.

\bibitem[Almeida et~al.(2021)Almeida, Astudillo, and Vandermeulen]{Almeida2021}
Diogo~F Almeida, Patricio Astudillo, and Dirk Vandermeulen.
\newblock Three-dimensional image volumes from two-dimensional digitally
  reconstructed radiographs: A deep learning approach in lower limb ct scans.
\newblock \emph{Medical Physics}, 48\penalty0 (5):\penalty0 2448--2457, 2021.

\bibitem[Bayat et~al.(2020)Bayat, Sekuboyina, Paetzold, Payer, Stern, Urschler,
  Kirschke, and Menze]{Bayat2020}
Amirhossein Bayat, Anjany Sekuboyina, Johannes~C Paetzold, Christian Payer,
  Darko Stern, Martin Urschler, Jan~S Kirschke, and Bjoern~H Menze.
\newblock Inferring the 3d standing spine posture from 2d radiographs.
\newblock In \emph{Medical Image Computing and Computer Assisted
  Intervention--MICCAI 2020: 23rd International Conference, Lima, Peru, October
  4--8, 2020, Proceedings, Part VI 23}, pages 775--784, 2020.

\bibitem[Nakao et~al.(2021)Nakao, Tong, Nakamura, and Matsuda]{Nakao2021}
Megumi Nakao, Fei Tong, Mitsuhiro Nakamura, and Tetsuya Matsuda.
\newblock Image-to-graph convolutional network for deformable shape
  reconstruction from a single projection image.
\newblock In \emph{Medical Image Computing and Computer Assisted
  Intervention--MICCAI 2021: 24th International Conference, Strasbourg, France,
  September 27--October 1, 2021, Proceedings, Part IV 24}, pages 259--268,
  2021.

\bibitem[Chen et~al.(2023)Chen, Guo, Zhang, Fang, He, and Wang]{Chen2023}
Zheye Chen, Lijun Guo, Rong Zhang, Zhongding Fang, Xiuchao He, and Jianhua
  Wang.
\newblock Bx2s-net: Learning to reconstruct 3d spinal structures from bi-planar
  x-ray images.
\newblock \emph{Computers in Biology and Medicine}, 154:\penalty0 106615, 2023.

\bibitem[Wu et~al.(2023)Wu, Zhang, Fang, Liu, Wang, Cui, and Shen]{Wu2023}
Han Wu, Jiadong Zhang, Yu~Fang, Zhentao Liu, Nizhuan Wang, Zhiming Cui, and
  Dinggang Shen.
\newblock Multi-view vertebra localization and identification from ct images.
\newblock In \emph{International Conference on Medical Image Computing and
  Computer-Assisted Intervention}, pages 136--145. Springer, 2023.

\bibitem[Liu et~al.(2021)Liu, Lin, Cao, Hu, Wei, Zhang, Lin, and Guo]{Liu2021}
Ze~Liu, Yutong Lin, Yue Cao, Han Hu, Yixuan Wei, Zheng Zhang, Stephen Lin, and
  Baining Guo.
\newblock Swin transformer: Hierarchical vision transformer using shifted
  windows.
\newblock In \emph{Proceedings of the IEEE/CVF international conference on
  computer vision}, pages 10012--10022, 2021.

\bibitem[Ronneberger et~al.(2015)Ronneberger, Fischer, and
  Brox]{Ronneberger2015}
Olaf Ronneberger, Philipp Fischer, and Thomas Brox.
\newblock U-net: Convolutional networks for biomedical image segmentation.
\newblock In \emph{Medical image computing and computer-assisted
  intervention--MICCAI 2015: 18th international conference, Munich, Germany,
  October 5-9, 2015, proceedings, part III 18}, pages 234--241, 2015.

\bibitem[Hatamizadeh et~al.(2021)Hatamizadeh, Nath, Tang, Yang, Roth, and
  Xu]{Hatamizadeh2021}
Ali Hatamizadeh, Vishwesh Nath, Yucheng Tang, Dong Yang, Holger~R Roth, and
  Daguang Xu.
\newblock Swin unetr: Swin transformers for semantic segmentation of brain
  tumors in mri images.
\newblock In \emph{International MICCAI Brainlesion Workshop}, pages 272--284,
  2021.

\bibitem[Wang et~al.(2023)Wang, Li, Mei, Wei, Liu, Wang, Sang, Yuille, Xie, and
  Zhou]{Wang2023}
Yiqing Wang, Zihan Li, Jieru Mei, Zihao Wei, Li~Liu, Chen Wang, Shengtian Sang,
  Alan~L Yuille, Cihang Xie, and Yuyin Zhou.
\newblock Swinmm: masked multi-view with swin transformers for 3d medical image
  segmentation.
\newblock In \emph{International Conference on Medical Image Computing and
  Computer-Assisted Intervention}, pages 486--496, 2023.

\bibitem[Tang et~al.(2022)Tang, Yang, Li, Roth, Landman, Xu, Nath, and
  Hatamizadeh]{Tang2022}
Yucheng Tang, Dong Yang, Wenqi Li, Holger~R Roth, Bennett Landman, Daguang Xu,
  Vishwesh Nath, and Ali Hatamizadeh.
\newblock Self-supervised pre-training of swin transformers for 3d medical
  image analysis.
\newblock In \emph{Proceedings of the IEEE/CVF conference on computer vision
  and pattern recognition}, pages 20730--20740, 2022.

\bibitem[Milletari et~al.(2016)Milletari, Navab, and Ahmadi]{Milletari2016}
Fausto Milletari, Nassir Navab, and Seyed-Ahmad Ahmadi.
\newblock V-net: Fully convolutional neural networks for volumetric medical
  image segmentation.
\newblock In \emph{2016 fourth international conference on 3D vision (3DV)},
  pages 565--571. Ieee, 2016.

\bibitem[Shannon(1948)]{Shannon1948}
Claude~Elwood Shannon.
\newblock A mathematical theory of communication.
\newblock \emph{The Bell system technical journal}, 27\penalty0 (3):\penalty0
  379--423, 1948.

\bibitem[Zhang et~al.(2018)Zhang, Xiang, Hospedales, and Lu]{Zhang2018}
Ying Zhang, Tao Xiang, Timothy~M Hospedales, and Huchuan Lu.
\newblock Deep mutual learning.
\newblock In \emph{Proceedings of the IEEE conference on computer vision and
  pattern recognition}, pages 4320--4328, 2018.

\bibitem[Milickovic et~al.(2000)Milickovic, Baltas, Giannouli, Lahanas, and
  Zamboglou]{Milickovic2000}
Natasa Milickovic, Dimos Baltas, S~Giannouli, M~Lahanas, and N~Zamboglou.
\newblock Ct imaging based digitally reconstructed radiographs and their
  application in brachytherapy.
\newblock \emph{Physics in Medicine \& Biology}, 45\penalty0 (10):\penalty0
  2787, 2000.

\bibitem[Wasserthal et~al.(2023)Wasserthal, Breit, Meyer, Pradella, Hinck,
  Sauter, Heye, Boll, Cyriac, Yang, et~al.]{Wasserthal2023}
Jakob Wasserthal, Hanns-Christian Breit, Manfred~T Meyer, Maurice Pradella,
  Daniel Hinck, Alexander~W Sauter, Tobias Heye, Daniel~T Boll, Joshy Cyriac,
  Shan Yang, et~al.
\newblock Totalsegmentator: Robust segmentation of 104 anatomic structures in
  ct images.
\newblock \emph{Radiology: Artificial Intelligence}, 5\penalty0 (5), 2023.

\bibitem[Deng et~al.(2021)Deng, Wang, Hui, Li, Li, Luo, Sun, Quan, Yang, Hao,
  et~al.]{Deng2021}
Yang Deng, Ce~Wang, Yuan Hui, Qian Li, Jun Li, Shiwei Luo, Mengke Sun, Quan
  Quan, Shuxin Yang, You Hao, et~al.
\newblock Ctspine1k: A large-scale dataset for spinal vertebrae segmentation in
  computed tomography.
\newblock \emph{arXiv preprint arXiv:2105.14711}, 2021.

\bibitem[Sekuboyina et~al.(2021)Sekuboyina, Husseini, Bayat, L{\"o}ffler,
  Liebl, Li, Tetteh, Kuka{\v{c}}ka, Payer, {\v{S}}tern, et~al.]{Sekuboyina2021}
Anjany Sekuboyina, Malek~E Husseini, Amirhossein Bayat, Maximilian L{\"o}ffler,
  Hans Liebl, Hongwei Li, Giles Tetteh, Jan Kuka{\v{c}}ka, Christian Payer,
  Darko {\v{S}}tern, et~al.
\newblock Verse: a vertebrae labelling and segmentation benchmark for
  multi-detector ct images.
\newblock \emph{Medical image analysis}, 73:\penalty0 102166, 2021.

\bibitem[Jin et~al.(2023)Jin, Gu, Wei, Adhinarta, Kuang, Zhang, Pfister, Ni,
  Yang, and Li]{Jin2023}
Liang Jin, Shixuan Gu, Donglai Wei, Jason~Ken Adhinarta, Kaiming Kuang,
  Yongjie~Jessica Zhang, Hanspeter Pfister, Bingbing Ni, Jiancheng Yang, and
  Ming Li.
\newblock Ribseg v2: A large-scale benchmark for rib labeling and anatomical
  centerline extraction.
\newblock \emph{IEEE Transactions on Medical Imaging}, 2023.

\bibitem[Kofler et~al.(2021)Kofler, Ezhov, Isensee, Balsiger, Berger, Koerner,
  Demiray, Rackerseder, Paetzold, Li, et~al.]{Kofler2021}
Florian Kofler, Ivan Ezhov, Fabian Isensee, Fabian Balsiger, Christoph Berger,
  Maximilian Koerner, Beatrice Demiray, Julia Rackerseder, Johannes Paetzold,
  Hongwei Li, et~al.
\newblock Are we using appropriate segmentation metrics? identifying correlates
  of human expert perception for cnn training beyond rolling the dice
  coefficient.
\newblock \emph{arXiv preprint arXiv:2103.06205}, 2021.

\bibitem[Cerveri et~al.(2010)Cerveri, Marchente, Bartels, Corten, Simon, and
  Manzotti]{Cerveri2010}
Pietro Cerveri, Mario Marchente, Ward Bartels, Kristoff Corten, Jean-Pierre
  Simon, and Alfonso Manzotti.
\newblock Automated method for computing the morphological and clinical
  parameters of the proximal femur using heuristic modeling techniques.
\newblock \emph{Annals of Biomedical Engineering}, 38:\penalty0 1752--1766,
  2010.

\bibitem[Fischer et~al.(2019)Fischer, Kroo{\ss}, Habor, and
  Radermacher]{Fischer2019}
Maximilian~CM Fischer, Felix Kroo{\ss}, Juliana Habor, and Klaus Radermacher.
\newblock A robust method for automatic identification of landmarks on surface
  models of the pelvis.
\newblock \emph{Scientific Reports}, 9\penalty0 (1):\penalty0 13322, 2019.

\bibitem[Di~Angelo and Di~Stefano(2015)]{Di2015}
Luca Di~Angelo and Paolo Di~Stefano.
\newblock A new method for the automatic identification of the dimensional
  features of vertebrae.
\newblock \emph{Computer methods and programs in biomedicine}, 121\penalty0
  (1):\penalty0 36--48, 2015.

\bibitem[Loshchilov and Hutter(2018)]{Loshchilov2018}
Ilya Loshchilov and Frank Hutter.
\newblock Decoupled weight decay regularization.
\newblock In \emph{International Conference on Learning Representations}, 2018.

\bibitem[Girdhar et~al.(2016)Girdhar, Fouhey, Rodriguez, and
  Gupta]{Girdhar2016}
Rohit Girdhar, David~F Fouhey, Mikel Rodriguez, and Abhinav Gupta.
\newblock Learning a predictable and generative vector representation for
  objects.
\newblock In \emph{Computer Vision--ECCV 2016: 14th European Conference,
  Amsterdam, the Netherlands, October 11-14, 2016, Proceedings, Part VI 14},
  pages 484--499. Springer, 2016.

\bibitem[Chen and Fang(2020)]{Chen2020}
Chih-Chia Chen and Yu-Hua Fang.
\newblock Using bi-planar x-ray images to reconstruct the spine structure by
  the convolution neural network.
\newblock In \emph{Future Trends in Biomedical and Health Informatics and
  Cybersecurity in Medical Devices: Proceedings of the International Conference
  on Biomedical and Health Informatics, ICBHI 2019, 17-20 April 2019, Taipei,
  Taiwan}, pages 80--85. Springer, 2020.

\bibitem[Hatamizadeh et~al.(2022)Hatamizadeh, Tang, Nath, Yang, Myronenko,
  Landman, Roth, and Xu]{Hatamizadeh2022}
Ali Hatamizadeh, Yucheng Tang, Vishwesh Nath, Dong Yang, Andriy Myronenko,
  Bennett Landman, Holger~R Roth, and Daguang Xu.
\newblock Unetr: Transformers for 3d medical image segmentation.
\newblock In \emph{Proceedings of the IEEE/CVF winter conference on
  applications of computer vision}, pages 574--584, 2022.

\bibitem[Oktay et~al.(2018)Oktay, Schlemper, Folgoc, Lee, Heinrich, Misawa,
  Mori, McDonagh, Hammerla, Kainz, et~al.]{Oktay2018}
Ozan Oktay, Jo~Schlemper, Loic~Le Folgoc, Matthew Lee, Mattias Heinrich,
  Kazunari Misawa, Kensaku Mori, Steven McDonagh, Nils~Y Hammerla, Bernhard
  Kainz, et~al.
\newblock Attention u-net: Learning where to look for the pancreas.
\newblock \emph{arXiv preprint arXiv:1804.03999}, 2018.

\end{thebibliography}

\end{document}